%% file: main.tex
\documentclass[10pt,logo,copyright]{nvidiatechreport}
\usepackage{xcolor}
\usepackage[numbers, sort]{natbib}
\definecolor{citecolor}{HTML}{0071BC}
\definecolor{linkcolor}{HTML}{ED1C24}

\renewcommand{\today}{2025-8-11}

\usepackage[utf8]{inputenc}
\usepackage[T1]{fontenc}
\usepackage{hyperref}
\usepackage{url}
\usepackage{booktabs}
\usepackage{amsfonts}
\usepackage{nicefrac}
\usepackage{microtype}

\input{preamble.tex}
\usepackage{capt-of}

\title{\MethodName: Video Pose Engine for 3D Geometric Perception}

\author{
Jiahui Huang, Qunjie Zhou, Hesam Rabeti, Aleksandr Korovko, Huan Ling, Xuanchi Ren, Tianchang Shen, Jun Gao, Dmitry Slepichev, Chen-Hsuan Lin, Jiawei Ren, Kevin Xie, Joydeep Biswas, Laura Leal-Taixe, Sanja Fidler
\\
NVIDIA\footnote{We acknowledge useful discussions from Aigul Dzhumamuratova, Viktor Kuznetsov, Soha Pouya, and Ming-Yu Liu, as well as release support from Vishal Kulkarni.}
\\
\url{https://research.nvidia.com/labs/toronto-ai/vipe/}
}

\input{sections/00-abstract.tex}

\begin{document}

\maketitle

\vspace{-2em}
\begin{figure}[htbp]
\centering
\includegraphics[width=\linewidth]{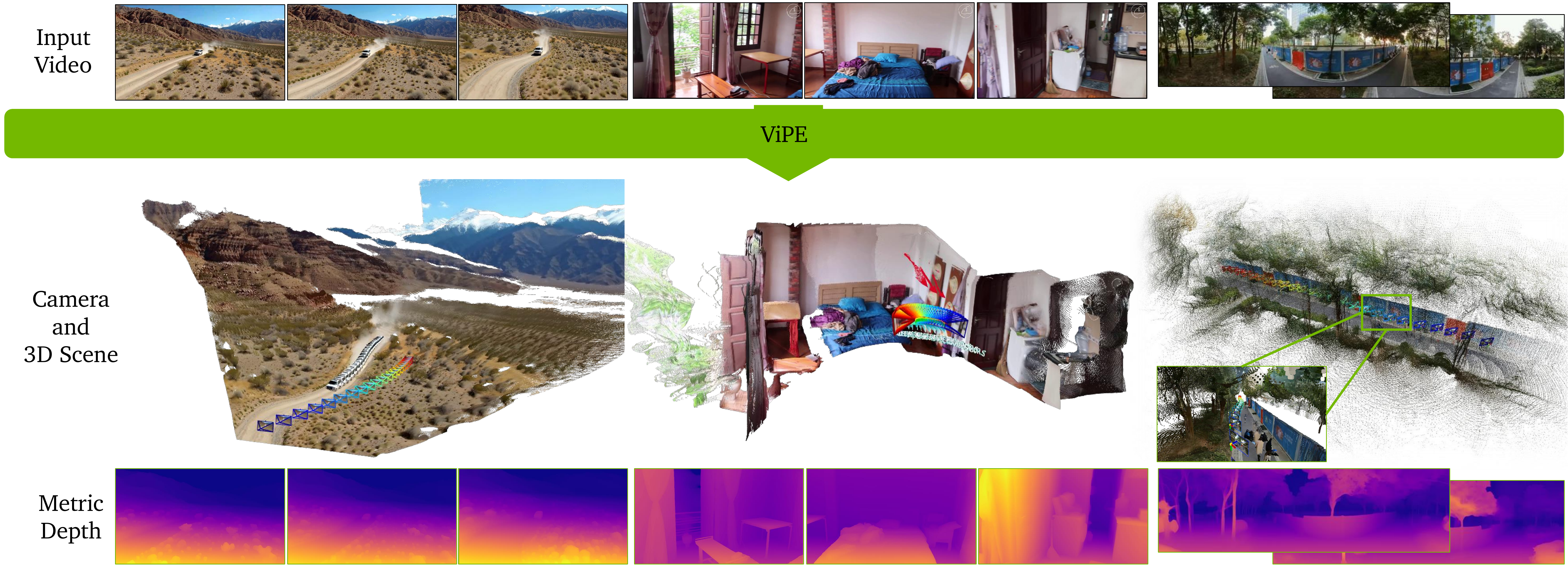}
\caption{We present \textbf{\MethodName}, a powerful and versatile video annotation engine. From a casual video, \MethodName outputs estimated camera motion and dense, metric-scale depth maps. \MethodName robustly handles diverse camera models, including standard perspective, wide-angle, or 360° panoramic videos.}
\label{fig:teaser}
\end{figure}

\abscontent

\input{sections/01-intro.tex}

\input{sections/02-related.tex}
\input{sections/03-method.tex}

\input{sections/04-exp.tex}

\input{sections/05-dataset.tex}
\input{sections/06-conclusion.tex}

\clearpage
{
\small

\bibliographystyle{abbrv}
\bibliography{main}

}

\end{document}

%% file: preamble.tex
\usepackage{epsfig}
\usepackage{graphicx}
\usepackage{amsmath,mathtools}
\usepackage{amssymb}
\usepackage{amsthm}
\usepackage{nicefrac}
\usepackage{microtype}
\usepackage{amsmath}
\usepackage{float}
\usepackage{amsfonts}
\usepackage{bm}
\usepackage{enumitem}
\usepackage{mathtools}
\usepackage{multirow}

\usepackage{fontawesome}
\usepackage{pifont}
\usepackage{color}
\usepackage[algo2e,inoutnumbered,linesnumbered,algoruled,vlined]{algorithm2e}
\usepackage{algpseudocode}
\usepackage{tabularx}
\usepackage{url}
\usepackage{bbm}
\usepackage{threeparttable}
\usepackage{tikz}
\usepackage{wrapfig}
\usetikzlibrary{backgrounds}
\usetikzlibrary{decorations.pathreplacing}
\usepackage[outline]{contour}

\newcommand{\MethodName}{ViPE\xspace}

\DeclareMathOperator*{\argmin}{argmin}

\newcommand{\PAR}[1]{\noindent{\bf #1~}}

\theoremstyle{definition}

\newcommand{\Pose}{{\mathbf{T}}}
\newcommand{\SE}{{\mathbb{SE}(3)}}
\newcommand{\intr}{{\bm{k}}}
\newcommand{\Depth}{{\mathbf{D}}}
\newcommand{\Graph}{{\mathcal{G}}}
\newcommand{\Vertex}{{\mathcal{V}}}
\newcommand{\Edge}{{\mathcal{E}}}
\newcommand{\uv}{{\mathbf{u}}}
\newcommand{\pts}{{\mathbf{p}}}
\newcommand{\Proj}{{\Pi}}
\newcommand{\Unproj}{\Pi^{-1}}
\newcommand{\Flow}{{\mathbf{F}}}
\newcommand{\Mask}{{\mathbf{M}}}

\newcommand{\RR}{\mathbb{R}}

\DeclarePairedDelimiterX{\infdivx}[2]{(}{)}{
  #1\;\delimsize\|\;#2
}

\definecolor{darkblue}{RGB}{49,130,189}
\definecolor{stanfordgrey}{RGB}{46,45,41}
\definecolor{cardinalred}{RGB}{253,141,60}

\usepackage[capitalize]{cleveref}

\crefname{section}{\S}{\S\S}
\crefname{subsection}{\S}{\S\S}
\crefname{conj}{Conj.}{Conj.}

\Crefname{assumption}{\textbf{H}\hspace{-3pt}}{\textbf{H}\hspace{-3pt}}
\crefname{assumption}{\textbf{H}}{\textbf{H}}

\crefname{algorithm}{\text{Alg.}}{\text{Alg.}}
\crefname{assumption}{\textbf{H}}{\textbf{H}}
\crefname{equation}{\text{Eq}}{\text{Eq}}
\crefname{definition}{\text{Dfn.}}{\text{Dfn.}}
\crefname{lemma}{\text{Lemma}}{\text{Lemma}}
\crefname{dfn}{\text{Dfn.}}{\text{Dfn.}}
\crefname{thm}{\text{Thm.}}{\text{Thm.}}
\crefname{tab}{\text{Tab.}}{\text{Tab.}}
\crefname{fig}{\text{Fig.}}{\text{Fig.}}
\crefname{table}{\text{Tab.}}{\text{Tab.}}
\crefname{figure}{\text{Fig.}}{\text{Fig.}}

\definecolor{mygreen}{RGB}{159, 200, 59}
\definecolor{myred}{RGB}{223, 135, 102}

\definecolor{first}{RGB}{118, 185, 0}
\definecolor{second}{RGB}{220,241,182}

%% file: sections/00-abstract.tex
\begin{abstract}

Accurate 3D geometric perception is an important prerequisite for a wide range of spatial AI systems. While state-of-the-art methods depend on large-scale training data, acquiring consistent and precise 3D annotations from in-the-wild videos remains a key challenge. In this work, we introduce \MethodName, a handy and versatile video processing engine designed to bridge this gap. \MethodName efficiently estimates camera intrinsics, camera motion, and dense, near-metric depth maps from unconstrained raw videos.
It is robust to diverse scenarios, including dynamic selfie videos, cinematic shots, or dashcams, and supports various camera models such as pinhole, wide-angle, and 360° panoramas.
We have benchmarked \MethodName on multiple benchmarks. Notably, it outperforms existing uncalibrated pose estimation baselines by 18\%/50\% on TUM/KITTI sequences, and runs at 3-5FPS on a single GPU for standard input resolutions.

We use \MethodName to annotate a large-scale collection of videos. This collection includes around 100K real-world internet videos, 1M high-quality AI-generated videos, and 2K panoramic videos, totaling approximately 96M frames -- all annotated with accurate camera poses and dense depth maps. We open-source \MethodName and the annotated dataset with the hope of accelerating the development of spatial AI systems.

\end{abstract}

%% file: sections/01-intro.tex
\section{Introduction}

The ability to understand 3D environments is a cornerstone of spatial intelligence for applications ranging from robotics to VR/AR, and autonomous systems. The foundational task of estimating low-level geometry—camera parameters and 3D scene structure remains a critical first step to many downstream technologies such as 3D reconstruction, camera or depth-conditioned video generation models, and training robotic policies.

Traditionally, this problem has been tackled by two main classes of methods. Classical Simultaneous Localization and Mapping (SLAM) systems~\cite{mur2015orb,davison2007monoslam} excel at estimating camera poses and sparse geometry from long video sequences, leveraging temporal consistency and loop closure~\cite{campos2021orb}. However, they typically assume a static scene and known camera intrinsics, and can be less robust to dynamic objects or degenerate motions. While some systems like COLMAP~\cite{schoenberger2016sfm} can refine intrinsics, jointly optimizing them with dense geometry for diverse, non-curated videos remains a challenge.

More recently, end-to-end feed-forward models~\cite{cong2025e3d} have emerged, trained on large datasets to directly regress camera poses and depth from images. While these methods show impressive robustness, their scalability is a significant bottleneck. Processing long videos is often intractable due to large GPU memory footprints, forcing practitioners to resort to subsampling video frames or processing short, disconnected chunks~\cite{prisacariu2017infinitam,maggio2025vggtslam}. A promising recent trend seeks a hybrid approach between SLAM and feed-forward approaches by integrating powerful learned front-ends like MaSt3R~\cite{leroy2024mast3r} into traditional SLAM back-ends, as demonstrated in systems like MASt3R-SLAM~\cite{murai2025mast3rslam} and concurrent work such as VGGT-SLAM~\cite{maggio2025vggtslam}. However, simply swapping the front-end is often insufficient in practice. As we demonstrate, these methods can still lack the accuracy and robustness required for large-scale annotation of diverse, in-the-wild videos, which motivates the need for a more tightly integrated system.

In this work, we introduce \textbf{Vi}deo \textbf{P}ose \textbf{E}ngine (shortened as \textbf{\MethodName}), designed to bridge the gap between classical and learning-based approaches. It combines the scalability and precision of a dense Bundle Adjustment (BA) framework, akin to SLAM, with the robustness of modern learned components. This synergy allows \MethodName to accurately and efficiently estimate camera poses, intrinsics, and dense, metric depth maps from challenging, in-the-wild videos.
Compared to the closest prior work, MegaSAM~\cite{li2025megasam}, \MethodName does not require per-frame optimization and is hence more efficient. More robust strategies are presented for handling dynamic objects, and a wider variety of camera models are supported. \MethodName is also experimentally shown to provide more accurate camera estimation results.
Speed-wise, \MethodName can typically reach a speed of \textbf{3-5FPS} on a single GPU\footnote{Measured with the input resolution of $640 \times 480$ on NVIDIA RTX 5090 GPU.}.
These advancements make our system uniquely suited for the demands of large-scale, diverse video annotation.

We put \MethodName in action to annotate a large-scale annotated dataset, which we publicly release along \MethodName. The dataset release comprises three distinct components: \textbf{Dynpose-100K++}, a re-annotation of approximately 100K challenging real-world internet videos with high-quality poses and dense geometry; \textbf{Wild-SDG-1M}, a large dataset of 1M high-quality, AI-generated videos sampled from video diffusion models; and \textbf{Web360}, a specialized dataset of annotated panoramic videos. Collectively, this release provides 96 million annotated frames across numerous varied sources, aiming to facilitate many downstream applications.

We summarize the key contributions:
\begin{itemize}[leftmargin=*, itemsep=0pt, topsep=0pt]
    \item A robust and efficient framework, \MethodName, for estimating camera parameters and dense depth from diverse, in-the-wild videos.
    \item A system design that integrates the strengths of classical SLAM (efficiency, scalability) and learned models (robustness), with key improvements over prior work in efficiency, dynamic object handling, and depth quality.
    \item A large-scale dataset of annotated videos, created using \MethodName, to facilitate future research in 3D computer vision.
\end{itemize}

%% file: sections/02-related.tex
\section{Related Works}
Our work builds upon decades of research in 3D reconstruction, as well as the recent paradigm shift towards using deep learning-based methods for low-level geometric tasks. We position our contributions relative to two main lines of work: classical optimization-based systems and modern feed-forward perception models.

\subsection{Visual SLAM and SfM}
The traditional approaches to 3D reconstruction from images can be divided into Structure-from-Motion (SfM) and Simultaneous Localization and Mapping (SLAM) techniques. SfM systems, such as the widely used COLMAP~\cite{schoenberger2016sfm,pan2024global}, are typically designed to process unordered collections of images, performing global optimization via Bundle Adjustment (BA) to recover highly accurate camera poses and sparse point clouds. SLAM systems, like ORB-SLAM~\cite{mur2015orb} and others~\cite{campos2021orb,davison2007monoslam,engel2017direct,korovko2025cuvslam}, are tailored for sequential video streams, processing frames incrementally to track camera motion in real-time while building a map of the environment.

Despite their success and precision, these classical methods face significant challenges with the ``in-the-wild" videos, which we focus on in our work. Their reliance on hand-crafted feature matching struggles in poorly-textured regions, and they are notoriously brittle in the presence of dynamic objects or non-rigid motion, which violates their fundamental static-world assumption. While systems like GLOMAP~\cite{pan2024global} have improved scalability, the core challenges remain. More recently, a trend towards purely data-driven SfM pipelines has emerged, where methods such as ParticleSfM~\cite{zhao2022particlesfm}, VGGSfM~\cite{wang2024vggsfm}, DATAP-SfM~\cite{ye2024datapsfm}, and DiffusionSfM~\cite{zhao2025diffusionsfm} leverage deep learning priors to replace classical components. While powerful, these approaches often focus on smaller-scale problems or specific aspects of the pipeline. \MethodName, on the other hand, targets a robust, scalable, and fully-integrated system.

\subsection{Feed-forward 3D Perception Models}

Classical methods, which depend on geometric consistency, often fail in challenging scenarios such as textureless regions, repetitive patterns, or wide-baseline views where feature matching is ambiguous. To address this, a recent wave of research has focused on feed-forward models that leverage powerful priors learned from large-scale datasets. This paradigm began with pairwise models like DUSt3R~\cite{wang2024dust3r} and MASt3R~\cite{leroy2024mast3r} and was quickly extended to multi-view settings that improved accuracy for general scenes~\cite{wang2024spann3r, cabon2025must3r, tang2025mvdust3r, yang2025fast3r, wang2025vggt,wang2025cut3r,lu2025align3r,xiao2025spatialtrackerv2,wang2025pi}. However, a critical bottleneck for this entire family of models is scalability; their computational and memory requirements grow quickly with the number of input frames, making it intractable to process long videos.

This limitation directly spurred the development of hybrid systems that integrate a robust feed-forward front-end into a classical SLAM~\cite{murai2025mast3rslam, maggio2025vggtslam, liu2025slam3r} or SfM~\cite{duisterhof2024mast3rsfm, elflein2025light3rsfm, liu2025regist3r} back-end to handle long sequences. While these hybrids represent a significant step, they often involve a ``loose coupling" that does not fully resolve inconsistencies between the learned front-end and the optimization back-end.

To further achieve high-quality, metric-scale dense geometry, our system also builds upon another important line of research: powerful monocular metric depth estimators~\cite{hu2024metric3d, piccinelli2025unik3d, piccinelli2025unidepthv2} and video depth models~\cite{chen2025videoda, chou2025flashdepth}. We integrate these as critical components at multiple stages: as a regularizing prior during optimization to resolve scale ambiguity, inspired by works like~\cite{wang2025depth}, and as a source for high-quality depth refinement.

In parallel, handling dynamic content remains a significant challenge. This has been tackled both by pure feed-forward models for pairwise image~\cite{zhang2024monst3r, chen2025easi3r, wang2024dust3r, sucar2025dynamicpointmap, zhang2025pomato, jin2024stereo4d,feng2025st4rtrack} or short video inputs~\cite{badki2025l4p,wimbauer2025anycam,jiang2025geo4d,xu2025geometrycrafter}, and by dense reconstruction systems like CasualSAM~\cite{zhang2022casualsam} and MegaSAM~\cite{li2025megasam}. Our work is most closely related to this latter category.
By addressing scalability, handling dynamic objects, and reaching metric scale in a unified framework, \MethodName introduces several key advantages over prior work, including a more efficient keyframe-based architecture, a more sophisticated strategy for modeling dynamics, and broader support for diverse camera models.

\subsection{Downstream Applications}
Large-scale datasets of videos annotated with accurate camera poses and 3D geometry are essential for a wide range of downstream applications. Such annotations serve both as valuable supervision signals during training and as informative inputs at test time. For example, they are currently widely used in training novel view synthesis methods, spanning diffusion-based models~\cite{zhou2025stable,gao2024cat3d,wu2025cat4d,ren2024scube,lu2024infinicube} and feed-forward reconstruction networks~\cite{liang2024feed,jin2024lvsm}. Camera trajectory information is also demonstrated to be useful in controllable video generation~\cite{ren2025gen3c,yu2025trajectorycrafter,ren2025cosmos,lu2024infinicube} tasks.
Additionally, accurate 3D geometric annotations can be helpful in deep-learning-based multi-view stereo (MVS) models~\cite{izquierdo2025mvsanywhere}, and is used for policy evaluation in embodied AI~\cite{team2025aether} and trajectory understanding~\cite{lin2025towards}.

A critical requirement across these applications is that the dataset must be large-scale, diverse in scene types, and contain high-quality geometric and pose annotations. Existing datasets~\cite{geiger2012we,sturm2012benchmark,greff2022kubric,baruch2021arkitscenes} often fall short in these aspects: many are small in scale and limited in diversity (e.g., focused only on indoor environments or constrained by fixed camera rigs). While other real-world datasets exist~\cite{jin2024stereo4d}, ours is the first to offer a combination of large-scale, diverse real-world content and high-quality annotations, making it uniquely suited for a broad spectrum of vision and robotics tasks.

%% file: sections/03-method.tex
\section{Methodology}

\begin{figure}[!tbp]
\includegraphics[width=\linewidth]{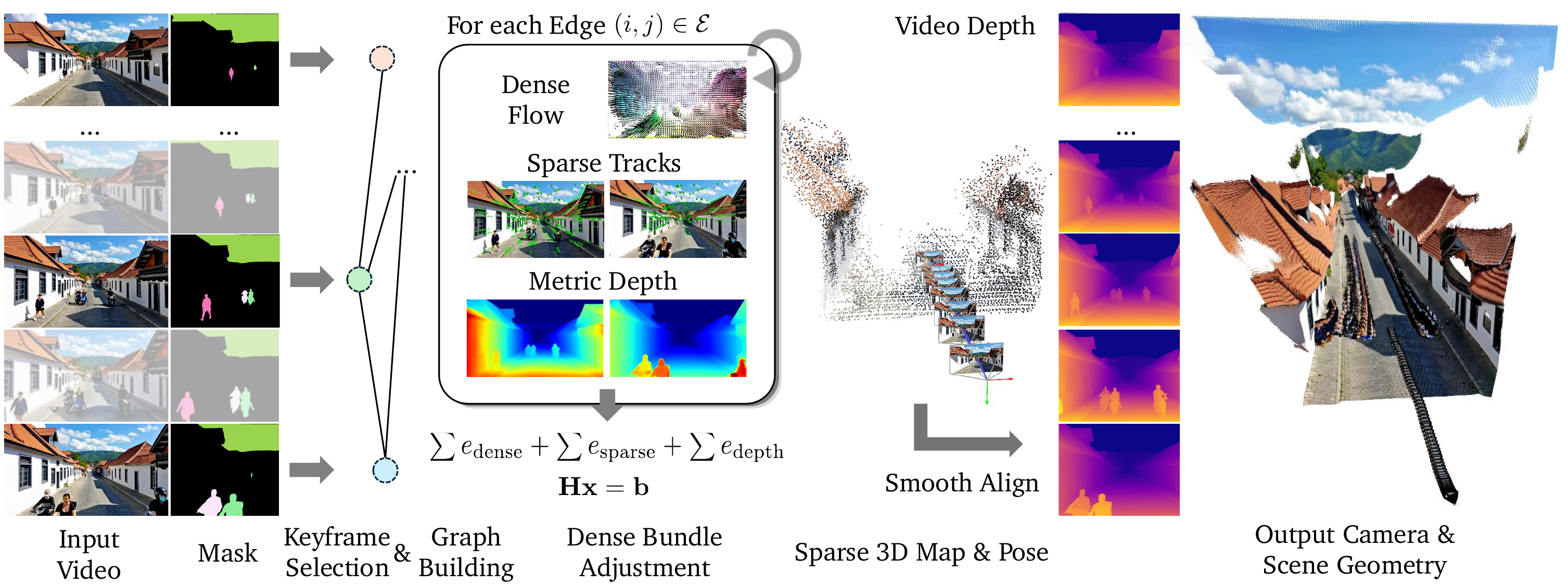}
\caption{\textbf{Pipeline} of \MethodName. The system takes a video as input and first estimates the semantic segmentation masks of the movable objects. It then estimates the camera poses, intrinsics, and depth maps from the video by solving a dense bundle adjustment problem incorporating various constraints. The final output is a dense depth map that is consistent with the camera poses and the intrinsics after the smooth depth alignment step.}
\label{fig:method}
\end{figure}

In this section, we present the core methodology of \MethodName, with an overview of the pipeline in \cref{sec:pipeline}, the core Bundle Adjustment (BA) formulation in \cref{sec:core}, and the details of the depth alignment stage to produce the final per-frame depth maps in \cref{subsec:pipeline:depth}.

\subsection{Overview}
\label{sec:pipeline}

The pipeline of \MethodName is based on a keyframe-based SLAM system for easier scalability and robustness to videos of arbitrary lengths.
It generally follows the same frontend and backend design as in most keyframe-based systems, primarily DROID-SLAM~\cite{teed2021droid}, consisting of the following steps (as illustrated in \cref{fig:method}):
\begin{enumerate}[leftmargin=*, itemsep=0pt, topsep=0pt]
  \item \textbf{Intrinsics Initialization:} An initial estimate of the camera intrinsics is obtained by uniformly sampling 4 frames from the video and running them through GeoCalib~\cite{veicht2024geocalib}.
  \item \textbf{Keyframe Selection:} For each incoming frame, we predict the motion from the current frame to the previous keyframe. The motion is a combination of the weighted optical flow from the dense flow network (\cref{subsec:core:dense}) and the sparse keypoint tracks (\cref{subsec:core:sparse}). If the motion is larger than a pre-defined threshold, we consider this a keyframe and add it to the BA graph $\Graph = (\Vertex, \Edge)$.
  \item \textbf{Frontend Tracking:} For the newly added keyframe, we build a graph $\Graph$ within a small sliding window of the last several keyframes. Frames within this window are connected by edges either if they are close in observation time step or if the co-visibility is high enough. The energy equation in \cref{sec:core} is then constructed and optimized using a Gauss-Newton solver.
  \item \textbf{Backend Optimization:} In the backend, a full BA optimization problem is solved that involves all the current keyframes, with the $\Graph$ built similarly as above. Camera intrinsic parameters are also unlocked for optimization at this stage. We empirically perform this optimization when the number of keyframes reaches 8, 16, and 64, as well as at the end of frontend tracking.
  \item \textbf{Pose Infilling:} Lastly, for each of the non-keyframes, we obtain the pose by building a small local graph that connects this frame to its closest two keyframes. We add uni-directional edges from the keyframe to the non-keyframe only, hence eliminating the need to compute the metric depth for all the frames during the BA optimization. The procedure is applied in parallel to all non-keyframes.
  \item \textbf{Dense Depth Estimation:} For each frame, we estimate a dense depth map in the same resolution as the input image that is consistent with the camera pose and the intrinsics. This process will be detailed in \cref{subsec:pipeline:depth}.
\end{enumerate}

\subsection{Formulation}
\label{sec:core}

At the core of \MethodName, we solve for a BA problem with the following unknown variables: frame poses $\{\Pose_i \in \SE\}$, camera intrinsics $\intr \in \RR^K$, and a low resolution depth map $\{\Depth_i \in \RR^{h \times w}\}$ for each keyframe $i$ in a graph $\Graph$:

\begin{equation}
  \label{eq:ba}
  e_\text{{\MethodName}}( \{\Pose_i\}, \{ \Depth_i \}, \intr ) = \sum_{(i,j) \in \Edge} e_\text{dense} (\Pose_i, \Pose_j, \Depth_i, \intr) + \sum_{(i,j) \in \Edge} e_\text{sparse} (\Pose_i, \Pose_j, \Depth_i, \intr) + \alpha \sum_{i\in \Vertex} e_\text{depth} ( \Depth_i ).
\end{equation}
Here $e_\text{dense}$ is the term that leverages dense matching between the two frames $i$ and $j$, while $e_\text{sparse}$ is a supplement term that leverages sparse keypoint matching information. $e_\text{depth}$ is a depth regularization term to ensure consistency and robustness of the pose estimation. These terms will be described in detail in \cref{subsec:core:dense,subsec:core:sparse,subsec:core:depth}.
The above energy function is minimized with respect to the unknown variables using a Gauss-Newton solver, and since the linear system is intrinsically sparse, we can efficiently solve it with a factorized solver using COLAMD reordering~\cite{davis2004algorithm}.
Since most of the casual videos contain dynamic objects and motions, we describe in \cref{subsec:pipeline:dynamic} how we mask the dynamic objects in the video.
In \cref{subsec:pipeline:camera}, we describe how the system supports different camera models.

\subsubsection{Dense Flow Constraint}
\label{subsec:core:dense}

The dense flow constraint is formulated as the same as in DROID-SLAM~\cite{teed2021droid}, where:
\begin{equation}
  \label{eq:dense}
  e_\text{dense} (\Pose_i, \Pose_j, \Depth_i, \intr) = \sum_{\uv} {w[\uv]} \cdot \big\lVert \Proj_\intr ( \Pose_j^{-1} \Pose_i \circ \Unproj_\intr ( \Depth_i [\uv] ) ) - \uv - \Flow_{ij}[\uv] \big\rVert^2.
\end{equation}
Here $\uv$ represents the pixel coordinates in the image, and the above equation is summed over all $h \times w$ existing pixels in the depth map $\Depth$.
We choose $h=H/8$ and $w=W/8$ to reduce the number of unknown variables to be optimized.
An additional optical flow $\Flow_{ij}$ is estimated between the two frames $i$ and $j$, which is regressed from an optical flow network from \cite{teed2021droid}.
Such an optical flow network takes two images as input and outputs a flow map $\Flow_{ij} \in \RR^{h \times w \times 2}$ in the same resolution as the depth map.
Internally the network builds a cost volume with an iterative refinement module, and provides a hint to the current estimation with a prior estimated flow $\Flow_{ij}^\text{prior} = \Proj_\intr ( \Pose_j^{-1} \Pose_i \circ \Unproj_\intr ( \Depth_i [\uv] ) )$ as the initial guidance of the cost volume.
In addition to the flow, a weight map $w[\uv]$ is also estimated (detailed in \cref{subsec:pipeline:dynamic}) to reflect the confidence of the flow estimation as well as the probability of motion, which is less useful for pose estimation.

\subsubsection{Sparse Point Constraint}
\label{subsec:core:sparse}

While the estimated dense optical flow $\Flow_{ij}$ is robust to various camera motions and texture-less scenes, due to its low resolution and network-inference nature, it might miss fine details visible only from the original high-resolution images that are critical for localization.
With this in mind, we propose a sparse point constraint based on an off-the-shelf CUDA-based fast feature detection and tracking module from the \texttt{cuVSLAM} package~\cite{korovko2025cuvslam}.
Internally, the features are generated by the Shi-Tomasi corner detector~\cite{shi1994good} and tracked using the Lucas-Kanade algorithm~\cite{lucas1981iterative}.
These features are computed on the original high-resolution image, providing sub-pixel constraints relative to the network's resolution, hence providing a physically grounded set of accurate flow vectors.
The sparse constraint is formulated as:
\begin{equation}
  \label{eq:sparse}
  e_\text{sparse} (\Pose_i, \Pose_j, \Depth_i, \intr) = \sum_{\pts_i} \big\lVert \Proj_\intr ( \Pose_j^{-1} \Pose_i \circ \Unproj_\intr ( \texttt{Bilerp}(\Depth_i, \pts_{i}) ) ) - \pts_{j} \big\rVert^2,
\end{equation}
where $\pts_i \in \RR^2$ and $\pts_j \in \RR^2$ are the matched sparse keypoints detected in frame $i$ and $j$, respectively, and $\texttt{Bilerp}(\Depth_i, \pts_{i})$ is the bilinear interpolation of the depth map $\Depth_i$ at the pixel coordinates $\pts_i$ with the prior assumption that the optimal depth map should be smoothly interpolated.

In practice, however, the above term would lead to a semi-sparse Hessian pattern when solving the BA problem since the Jacobian of one $e_\text{sparse}$ term is related to multiple (up to 4 neighbourhood) pixel locations in $\Depth$, creating numerous interactions between the depth maps in the graph $\Graph$ themselves.
Although a highly efficient solver is proposed in prior works such as \cite{huang2023neural}, we found it more efficient and effective to use a simpler constraint by replacing the bilinear interpolation with a bilinear splatting operation, yielding the same constraint as in \cref{eq:dense} but replacing its depth map $\Depth$ with $\texttt{Bisplat} (\{\pts_j - \pts_i \}, \{ \pts_i\})$,
where \texttt{Bisplat} is the bilinear splatting operation that assigns each pixel location $\uv$ an accumulated value weighted by the distance to all the input locations $\pts_i$.

\subsubsection{Depth Regularization}
\label{subsec:core:depth}
Similarly to the flow correspondences, accurate depth map estimations are typically crucial for resolving ambiguities, especially for small (or degenerate) camera motions.
We hence add a depth regularization term as:
\begin{equation}
  \label{eq:depth}
  e_\text{depth} (\Depth_i) = \sum_{\uv} m[\uv] \cdot \big\lVert \Depth_i[\uv] - \Depth_i^\text{prior}[\uv] \big\rVert^2,
\end{equation}
where $\Depth_i^\text{prior}$ is the prior depth map estimated from a pre-trained monocular metric depth estimation network (with $m$ being the estimation uncertainty).
We allow the users to choose from different depth estimation models, including Metric3dv2~\cite{hu2024metric3d}, UniDepthV2~\cite{piccinelli2025unidepthv2}, as well as UniK3D~\cite{piccinelli2025unik3d}, depending on the camera models\footnote{In all of our quantitative experiments, we choose Metric3dv2~\cite{hu2024metric3d} consistently.}.
All these networks are based on single images and provide an estimate of the current scene scale, hence not only help reduce the scale drifting issue commonly seen in SLAM systems, but also provide a good estimate of the real-world metric scale.

Notably, for the metric depth estimation models, the predicted depth maps are conditioned on the camera intrinsics $\intr$.
We hence update the depth predictions after the intrinsics are optimized, and replace the prior depth maps $\Depth_i^\text{prior}$ with the newly predicted ones (for \cite{hu2024metric3d} this is a simple scaling operation).

\subsubsection{Dynamic Object Masking}
\label{subsec:pipeline:dynamic}

Many real-world videos usually have dynamic objects occupying a large pixel region of the video, revealing challenging ambiguities determining the static background that camera poses $\Pose$ depend on.
Most recent state-of-the-art motion segmentation methods (such as \cite{goli2024romo,huang2025segment}) combine semantic priors and optical flow for robustly segmenting the dynamic objects.
We hereby take a simpler approach by using pure semantic segmentation information for its robustness and efficiency.
Specifically, given a list of user-specified semantic classes, we follow SAM-Track~\cite{cheng2023segment}, and first apply GroundingDINO~\cite{liu2024grounding} to provide bounding box prompts to the Segment Anything~\cite{kirillov2023segment} model, obtaining the segmentation masks of these classes.
Instead of applying the above two models on each frame which is computationally expensive, we apply them at a fixed frame interval and propagate the segmentation masks through XMem~\cite{cheng2022xmem}.

The output of the semantic masks is then inverted to obtain the static background mask as $\Mask$.
For \cref{eq:dense}, we multiply $\Mask$ with the weight regressed by the dense flow network to obtain the weight map $w$.
Similarly, for \cref{eq:sparse}, we remove all the point tracks outside the region of $\Mask$ and treat them as outliers.

\subsubsection{Handling Different Camera Models}
\label{subsec:pipeline:camera}

\MethodName provides support for various camera models as well as optimization of their intrinsic parameters $\intr$.
Our system is based on the assumption of the following radial camera formulation:
\begin{equation}
  \uv = \Proj_\intr \left(\begin{bmatrix} x \\ y \\ z \end{bmatrix}\right) = \begin{bmatrix} f \cdot q_\intr(\theta) \cdot \cos \phi + W/2 \\ f \cdot q_\intr (\theta) \cdot \sin \phi + H/2 \end{bmatrix},
\end{equation}
where $x,y,z$ are 3D coordinates in the camera local space, $\theta = \arctan \frac{\sqrt{x^2 + y^2}}{z}$ is the angle between the corresponding ray and the optical axis, and $\phi = \arctan \frac{y}{x}$ is the rotation angle of the projected point on the canvas.
Note that for simplicity we always assume that the principal point is fixed at the image center $(\frac{W}{2}, \frac{H}{2})$, and the focal length $f$ is the same for both axes.

\begin{figure}
\centering
\includegraphics[width=\linewidth]{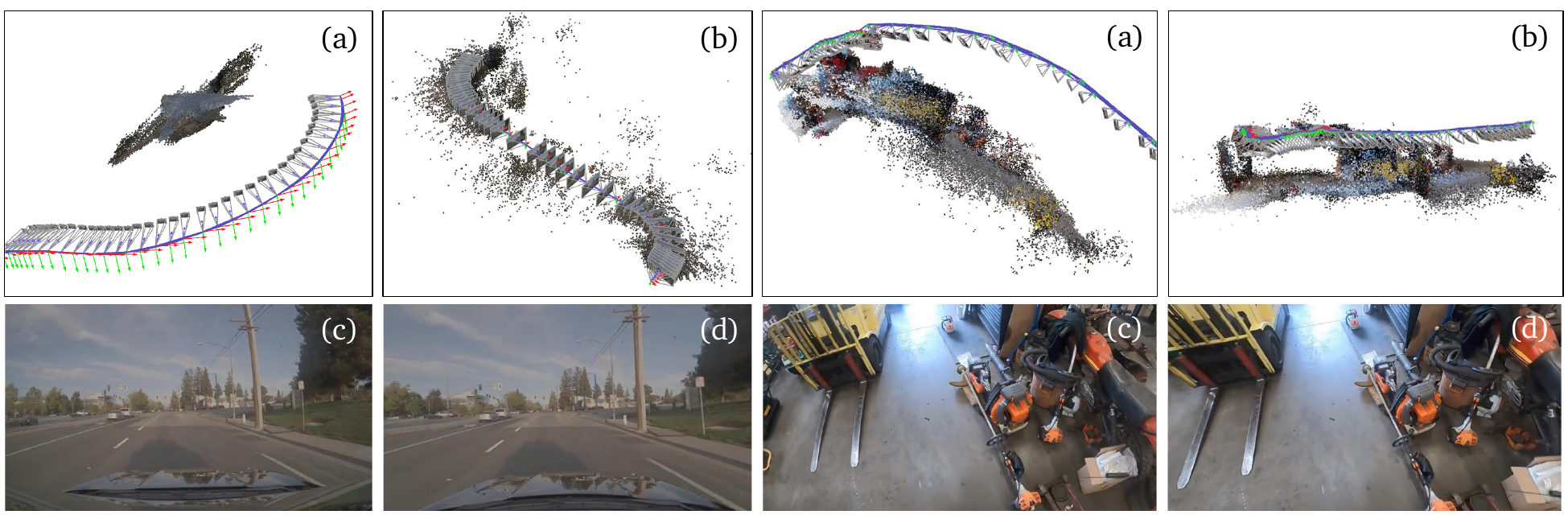}
\caption{Pose estimation results on \textbf{wide-angle cameras}. (a) Baseline results with the pinhole camera assumption. (b) \MethodName's results using the unified camera model. (c) Sample frames from the video. (d) Rectified frames using \MethodName's intrinsics estimation.}
\label{fig:qual_mei}
\end{figure}

For a simple pinhole camera, we let $q_\intr (\theta) = \tan \theta$, and $f$ is the only scalar parameter to be estimated in $\intr$.
For wide-angle/fisheye camera, we follow \cite{hagemann2023deep} and use the unified camera model~\cite{mei2007single} where $q_\intr(\theta) = \frac{\tan \theta}{1 + \alpha \sqrt{\tan^2 \theta + 1}}$ and $\intr = [f, \alpha] \in \RR^2$. Here $\alpha$ controls the distortion strength, with $\alpha=0$ falling back to the pinhole camera model.
\cref{fig:qual_mei} shows examples of the pose estimation results on videos captured by wide-angle cameras.

\MethodName's BA-based formulation naturally extends to multi-camera rigs. This is achieved by expanding the transformation $\Pose_i$ in \cref{eq:ba} to $\Pose_v \Pose_i$, where $\Pose_v$ is the transformation from the rig to the $v$-th camera's reference frame.
In \cite{teed2021droid}, the two cameras on a stereo rig are correlated by adding an additional set of edges in the graph that connects the left and right cameras at the same time step. This might become less effective in finding co-visible landmarks if the view frustums of the cameras have little overlap. We hence adaptively add cross-view edges in the graph based on the co-visibility of the cameras measured by projecting the dense depth maps to the other cameras' views.
360 camera is a special case of such a setting where the panorama image is stitched from two or more fisheye cameras.
To tackle these videos, we project the original image into 6 pinhole cameras (facing towards front, back, left, right, up, and bottom directions) roughly covering the cubical surface, and fix the relative transformation $\Pose_v$ during the optimization.

\subsection{Post-processed Dense Depth Alignment}
\label{subsec:pipeline:depth}

While state-of-the-art dense depth estimation networks~\cite{chen2025videoda,yang2024depth} can produce high-quality relative depth maps, it is challenging to recover a consistent absolute scale with the estimated camera pose $\{ \Pose_i \}$ across the entire video sequence.
On the other hand, the dense depth map $\Depth$ solved from the bundle adjustment~\cref{eq:ba} typically has a better alignment with the camera poses, and shows consistency across the entire video sequence. They can be, however, noisy/incomplete (especially in textureless regions) and suffer from low resolutions.

To achieve the best of both worlds, we propose a smooth depth alignment strategy.
We first use a video depth estimation network~\cite{chen2025videoda} to estimate a temporally smooth yet affine-invariant depth map for each frame $i$, denoted as $\Depth_i^\text{VDA} \in \RR^{H \times W}$.
In parallel we aggregate the point cloud unprojected with $\{\Depth_i\}$ from the BA optimization, filter the pixels that fail the consistency check with the estimated camera poses, and project them back to the image space to obtain a sparse depth map $\Depth_i^\text{BA} \in \RR^{H \times W}$.
We then make use  of a momentum-based update strategy to find the best affine transformation parameters:
\begin{equation}
  \begin{aligned}
    \alpha_i, \beta_i = \argmin_{\alpha, \beta} \sum_{\uv \text{ is valid}} \left\lVert \Mask \cdot (\alpha /\Depth_i^\text{VDA}[\uv] + \beta - 1/\Depth_i^\text{BA}[\uv]) \right\rVert_2^2, \\
    \hat{\alpha}_i = m \cdot \hat{\alpha}_{i-1} + (1-m) \cdot \alpha_i, \quad \hat{\beta}_i = m \cdot \hat{\beta}_{i-1} + (1-m) \cdot \beta_i,
  \end{aligned}
\end{equation}
where $m$ is the momentum factor, and the final depth map we output is $\Depth_i^\text{HD} = \frac{1}{\hat{\alpha}_i / \Depth_i^\text{VDA} + \hat{\beta}_i}$.

Notably, real-world videos can be diverse in terms of the scene distributions, and the projected depth map $\Depth^\text{BA}_i$ might not have enough information to constrain the above affine transformation.
We hence compute the percentage of the pixels that are covered by this depth map, and apply PriorDA~\cite{wang2025depth} to infill the depth map conditioned on the partial observation as well as the input image before assigning this to $\Depth_i^\text{BA}$ for alignment.
Under very extreme and rare cases where only a few of or none of the pixels are covered, we directly assign $\Depth_i^\text{BA}$ to be the metric depth estimation from \cref{subsec:core:depth}.

%% file: sections/04-exp.tex
\section{Evaluation}

To fully demonstrate the capability of \MethodName, we compare ourselves to the state-of-the-art methods on the fundamental geometry estimation tasks, including camera intrinsics, poses, and depth estimation across standard benchmarks as well as \textit{in-the-wild} casual videos.

\subsection{Camera Pose Estimation}

\label{subsec:pose}

\subsubsection{Evaluation on Standard Benchmarks}
\label{subsec:pose_standard}

We first demonstrate our competitive performance by evaluating against established baselines on widely recognized datasets with readily available ground truth.

\PAR{Datasets.}
We measure the accuracy of the estimated camera pose and intrinsics on two main scenarios: (1) Indoor scenes represented by the widely used \textbf{TUM RGB-D} dataset~\cite{sturm2012benchmark} with multiple loop closures and complicated camera trajectory with scene motions; (2) Outdoor driving scenes.
For the latter, we crop the ultra-wide images from the \textbf{KITTI} odometry~\cite{geiger2012we} dataset to a resolution of $512 \times 368$ and only keep the first 1024 frames for simplicity.
Since all the sequences from the KITTI dataset are captured with a fixed camera intrinsics, to further evaluate the robustness of our method to different camera focal lengths, we supplement with an additional set \textbf{RDS}. This dataset consists of a subset of 64 sequences sampled from the original Real Driving Scene (RDS) dataset~\cite{agarwal2025cosmos,alhaija2025cosmos}.
We resample the input images to a pinhole with varying focal lengths ranging from 30 to 70 degrees.
This further tests the robustness of \MethodName on more complicated driving scenarios.

\PAR{Baselines and metrics.}
We compare to various state-of-the-art baselines available in the literature at the time of release. Out of them, MASt3R-SLAM~\cite{murai2025mast3rslam}, VGGT~\cite{wang2025vggt} and MegaSAM~\cite{li2025megasam} allow raw video input.
For a fair comparison, we executed MASt3R-SLAM in its uncalibrated configuration to acquire camera poses for all frames, mirroring the approach taken by other baselines. Additionally, for comprehensive reference, we present MASt3R-SLAM's metrics, specifically computed on keyframes (MASt3R-SLAM$^\text{KF}$), in \cref{table:driving_pose}, adhering to the methodology outlined in its original publication.
To facilitate VGGT inference on videos comprising hundreds to thousands of frames, we devised a sliding window strategy. This approach involves running VGGT on each local window, defined by $N$frames, with a guaranteed overlap of $K$ frames between successive windows. Subsequently, we estimate the similarity transformation to align the adjacent window predictions. This estimation leverages point maps accumulated from the overlapping frames, where we select the top $50\%$ of points based on their confidence scores predicted by VGGT. Our experimental observations suggest that employing larger local window sizes for VGGT generally leads to more accurate results, primarily by reducing error propagation from Sim(3) alignment on dense point maps. Through empirical evaluation, we determined that $N=120/200$ and $K=5$ provided the optimal performance when inference was performed on a single GPU.
Due to its simplicity and efficiency, we also add DROID-SLAM~\cite{teed2021droid} as a reference baseline, where the camera intrinsics is directly estimated via GeoCalib~\cite{veicht2024geocalib} using the first 2s of video (denoted as DROID-SLAM$^\dagger$).

We compute Absolute Trajectory Error (ATE), Relative Translation/Rotation Error (RTE, RRE), and pinhole intrinsics error (Focal).
ATE, RTE, RRE are classical SLAM metrics that measure how the predicted pose deviates from the ground truth after optimal rigid alignment, reflecting the pose quality in terms of both global and local scales~\cite{mur2015orb,sturm2012benchmark}.
The intrinsics error is computed as the absolute difference between prediction and ground-truth field of view angles.

\begin{table}[!t]
\footnotesize
\vspace{-1em}
\begin{center}
\setlength{\tabcolsep}{3.4pt}
\begin{tabular}{l|cccc|cccc|c}
\toprule
            & \multicolumn{4}{c|}{\texttt{Freiburg1} (static)}                                                                                   & \multicolumn{4}{c|}{\texttt{Freiburg3} (dynamic)}                                      & \multicolumn{1}{c}{\multirow{2}{*}{\begin{tabular}[c]{@{}c@{}}Run\\ Time$^\dagger$\end{tabular}}}                                           \\
            & \multicolumn{1}{c}{ATE (cm) $\downarrow$} & \multicolumn{1}{c}{RTE (cm) $\downarrow$} & \multicolumn{1}{c}{RRE ($^\circ$) $\downarrow$} & \multicolumn{1}{c|}{Focal ($^\circ$) $\downarrow$} & \multicolumn{1}{c}{ATE (cm) $\downarrow$} & \multicolumn{1}{c}{RTE (cm) $\downarrow$} & \multicolumn{1}{c}{RRE ($^\circ$) $\downarrow$} & \multicolumn{1}{c|}{Focal ($^\circ$) $\downarrow$} & \\ \midrule
DROID-SLAM$^\dagger$~\cite{teed2021droid}  & \cellcolor{second}{4.4} & \cellcolor{first}{0.6}  & \cellcolor{second}{0.39}   & \cellcolor{second}{4.1}  & \cellcolor{second}{2.7} & 1.0 &  \cellcolor{second}{0.27}  &  \cellcolor{second}{4.3}  & $\sim$2min \\
MASt3R-SLAM~\cite{murai2025mast3rslam} & 6.8  &  2.3 & 0.54 & N/A & 7.6 & 2.7 & 0.41 & N/A & \cellcolor{first}{$\sim$1.5min} \\
VGGT~\cite{wang2025vggt}        & 8.4  & 0.8   & 0.44   & 11.1   & 12.9   & \cellcolor{first}{0.5}  & 0.30 & 10.1  & $\sim$4min \\
MegaSAM~\cite{li2025megasam}     & 7.0 & \cellcolor{first}{0.6} & \cellcolor{first}{0.37}  &  10.5 & \cellcolor{first}{1.5} & \cellcolor{second}{0.8} & \cellcolor{first}{0.26} & 12.6 & $\sim$15min \\ \midrule
\textbf{Ours}  &   \cellcolor{first}{3.6}  &  \cellcolor{second}{0.7}  &    \cellcolor{second}{0.39}  &   \cellcolor{first}{1.8} &  \cellcolor{first}{1.5} &  \cellcolor{second}{0.8} &  \cellcolor{second}{0.27}  & \cellcolor{first}{0.6} & \cellcolor{second}{$\sim$3min} \\ \bottomrule
\end{tabular}
\end{center}
\vspace{-1em}
$^\dagger$: Run time is measured on a single NVIDIA RTX 5090 GPU, excluding the per-frame dense depth estimation time if possible. The dataset has $\sim$950 frames per sequence on average.
\vspace{-1em}
\caption{Pose and intrinsics accuracy measured on \textbf{TUM-RGBD}~\cite{sturm2012benchmark} dataset.}
\label{table:tum_pose}
\end{table}

\begin{table}[!t]
\footnotesize

\begin{center}
\setlength{\tabcolsep}{10.0pt}
\begin{tabular}{l|cc|cc|c}
\toprule
& \multicolumn{2}{c|}{\textbf{KITTI}~\cite{geiger2012we}}  & \multicolumn{2}{c|}{\textbf{RDS}~\cite{agarwal2025cosmos}}  & \multirow{2}{*}{\begin{tabular}[c]{@{}c@{}}Run\\ Time\end{tabular}} \\
& \multicolumn{1}{c}{ATE (m) $\downarrow$} & \multicolumn{1}{c|}{Focal ($^\circ$) $\downarrow$} & \multicolumn{1}{c}{ATE (m) $\downarrow$} & \multicolumn{1}{c|}{Focal ($^\circ$) $\downarrow$} & \\ \midrule
MASt3R-SLAM~\cite{murai2025mast3rslam}  & 122.2 & N/A & 21.0  & N/A & \cellcolor{second}{$\sim$4min} \\
MASt3R-SLAM$^\text{KF}$~\cite{murai2025mast3rslam}  & \cellcolor{second}{21.3} & N/A & 9.5 & N/A & - \\
VGGT~\cite{wang2025vggt}  & 23.8 & \cellcolor{first}{1.9} & \cellcolor{second}{5.7}    & \cellcolor{first}{5.9} & \cellcolor{first}{$\sim$3min} \\
MegaSAM~\cite{li2025megasam}     & 25.4 & \cellcolor{second}{2.3} & 9.3  & 47.7 & $\sim$17min \\ \midrule
\textbf{Ours}  & \cellcolor{first}{9.2}  &  \cellcolor{first}{1.9}  & \cellcolor{first}{5.0}  &  \cellcolor{second}{7.9} & $\sim$4.5min \\ \bottomrule
\end{tabular}
\end{center}
\vspace{-1.5em}
\caption{Pose and intrinsics accuracy measured on \textbf{outdoor driving} datasets.}
\label{table:driving_pose}
\end{table}

\begin{figure}[!t]
\centering
\includegraphics[width=0.48\linewidth]{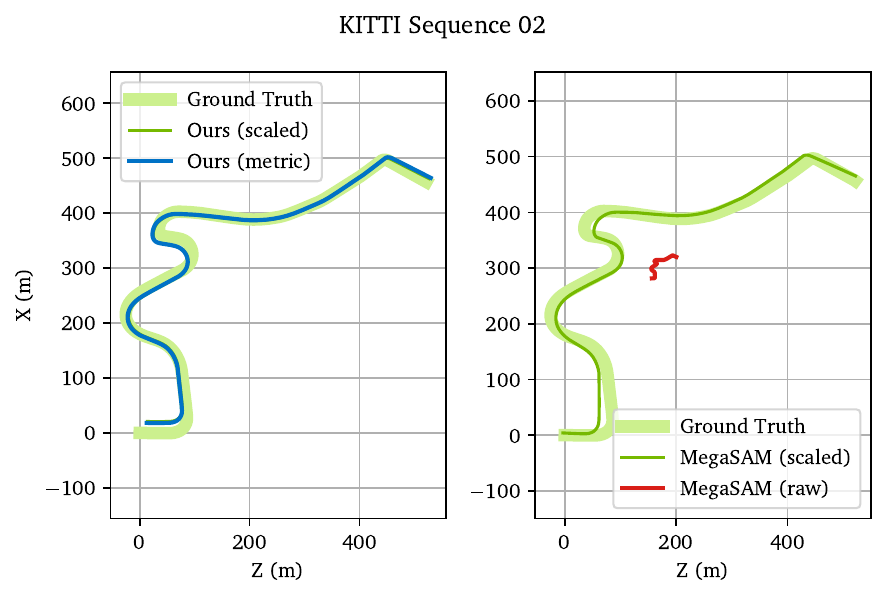}
\includegraphics[width=0.48\linewidth]{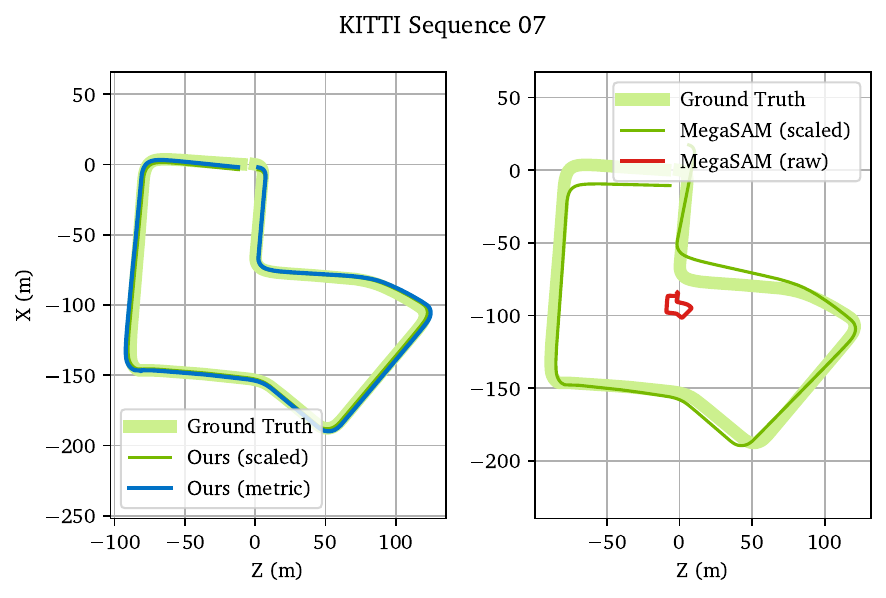}
\caption{Qualitative results of camera pose estimation on \textbf{KITTI dataset}~\cite{geiger2012we}. Output of \MethodName can be used as an approximation of the metric scale in real world, while the baseline \cite{li2025megasam} is not guaranteed to be scale-consistent.}
\label{fig:kitti_pose}
\end{figure}

\PAR{Results.}
As quantitatively shown in \cref{table:tum_pose,table:driving_pose}, \MethodName reaches competitive performance in both indoor and outdoor datasets.
The method is also robust to dynamic scenes by properly and efficiently removing movable objects from the camera estimation.
Notably, the scale of the output pose is roughly in line with the real-world scale thanks to the metric depth prediction module, as demonstrated in \cref{fig:kitti_pose}, where the baseline, \eg MegaSAM, outputs pose in an indefinite scale space.

\subsubsection{Evaluation on Unposed Videos}
\label{subsec:pose_unposed}

\PAR{Camera pose consistency metrics.}
In addition to the common evaluation setup, we further showcase the applicability of \MethodName for real-world \textit{in-the-wild} videos, where we are not equipped with ground-truth poses for evaluation. Hence we propose two new metrics for camera pose evaluation without ground-truth annotations.

\begin{itemize}[leftmargin=*, itemsep=0pt, topsep=0pt]
    \item \textbf{Shuttle Pose Error.} We feed both the video itself and a reversed version of it into the algorithm, obtaining two independent trajectories. Pose (ATE, RRE) and calibration errors (Focal) are then measured on these two sets of estimations, which we denote as S-ATE, S-RTE, and S-Focal. To make sure that the metric numbers are comparable across different baselines, we normalize the lengths of the trajectories to 1 before computing the rigid alignment between them.
    \item \textbf{Sampson Error.} We define the Sampson error as the first-order approximation of the distance from one interest point detected by LightGlue~\cite{lindenberger2023lightglue} to its corresponding epipolar line in the subsequent frame: \begin{align}
        \frac{1}{N} \sum_{i=1}^{N-1} \frac{1}{K} \sum_{k=1}^{K} \frac{|\bar{\bm{y}}_{ik}^\top \mathbf{F} \bar{\bm{x}}_{ik}|}{\sqrt{\left\|\mathbf{S}\mathbf{F} \bar{\bm{x}}_{ik}\right\|_2^2 + \left\|\mathbf{S}\mathbf{F}^\top \bar{\bm{y}}_{ik}\right\|_2^2}} \; , \quad \text{where} \; \mathbf{S} = \begin{bmatrix}
            1 & 0 & 0 \\
            0 & 1 & 0 \\
            0 & 0 & 0
        \end{bmatrix}.
    \end{align}
    Here $N$ it the total number of frames and $K$ is the number of correspondences. The correspondences themselves between frame $i$ and frame $i+1$, \ie $\bar{\bm{x}}_{ik}$ and $\bar{\bm{y}}_{ik}$, are expressed as homogeneous coordinates. The fundamental matrix $\mathbf{F}$ is computed from the output pinhole intrinsic parameters.
\end{itemize}

\begin{wrapfigure}{r}{0.4\textwidth}
\centering
\vspace{-10pt}
\includegraphics[width=0.37\textwidth]{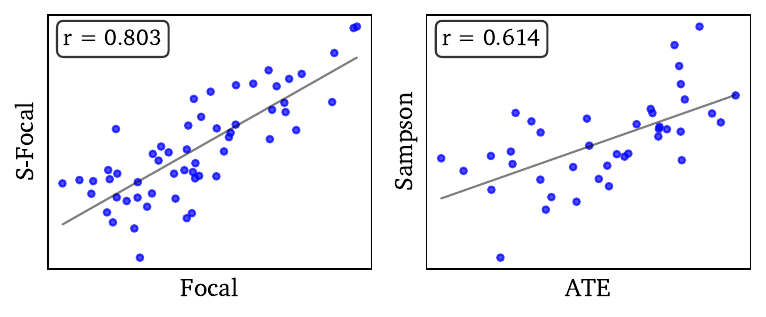}
\vspace{-20pt}
\end{wrapfigure}
To demonstrate the effectiveness of the above-proposed metrics, we utilize a sub-dataset from \cref{subsec:pose_standard} and compute both the consistency metrics and the standard metrics. As shown in the inset, the proposed focal and pose errors are generally correlated to the standard pose errors (with $r \in (-1,1)$ denoting the Pearson correlation coefficient).

\PAR{Dataset.}
We gather two subsets of datasets for benchmarking: (1) \textbf{OpenDV} dataset is a random subset of 50 videos from \cite{yang2024generalized}. The dataset contains mainly dashcam videos mounted on a driving vehicle recording the road and various surrounding environments around the globe. These videos are demonstrated to benefit self-driving applications. (2) \textbf{VidBench} dataset is a random subset of 60 videos from the Dynpose-100K dataset~\cite{rockwell2025dynamic}, which contains videos gathered from the web, including various scenes recorded either by hand-held cameras or professional ones. The videos are diverse in terms of motions and scene distribution, covering both indoor and outdoor scenes.

\begin{figure}[!t]
\centering
\includegraphics[width=\linewidth]{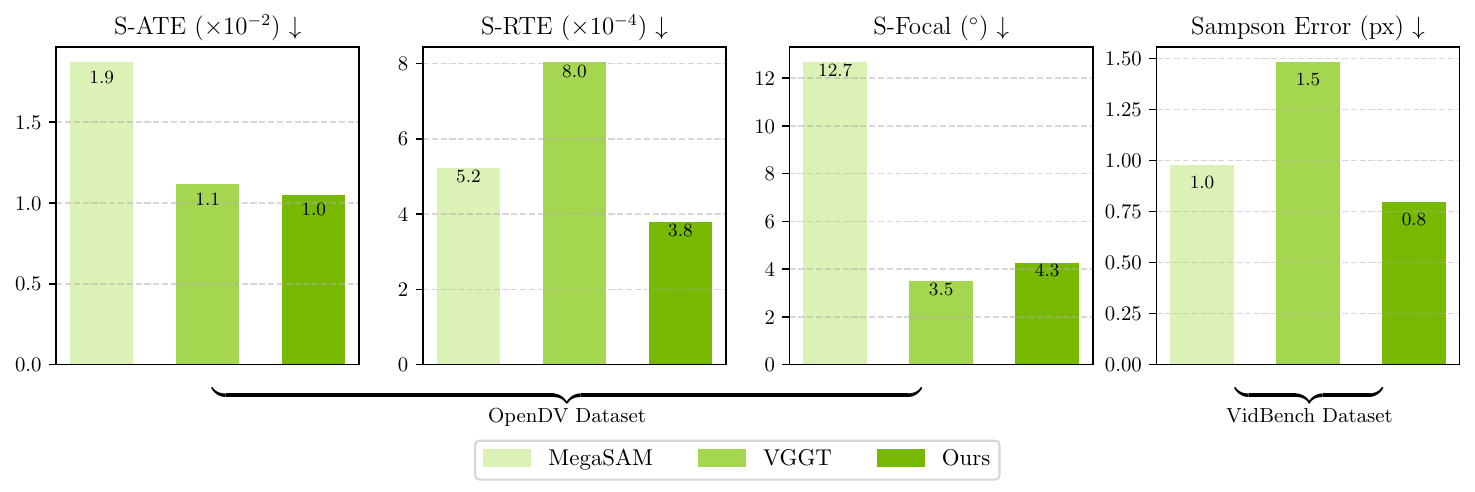}
\caption{Qualitative results of camera pose estimation on \textbf{unposed videos} using the proposed metrics.}
\label{fig:wild_comparison}
\end{figure}

\PAR{Results.} As shown in \cref{fig:wild_comparison}, our method reaches better consistency in the shuttle measurement and lower Sampson error, indicating that the estimated camera poses are more reliable. This demonstrates the wide applicability of \MethodName in real-world scenarios.

\subsection{Depth Estimation}
\label{subsec:depth}

\PAR{Datasets.} We evaluate \MethodName for depth estimation on  the two well-established benchmarks, \ie,
MPI-Sintel~\cite{butler2012naturalistic} (\textbf{SINTEL}) dataset and the \textbf{ETH3D}~\cite{schops2019bad} dataset.
For the SINTEL synthetic dataset, we manually select 6 representative sequences\footnote{Sequences are \texttt{alley\_2}, \texttt{bamboo\_1}, \texttt{bamboo\_2}, \texttt{sleeping\_1}, \texttt{sleeping\_2}, \texttt{temple\_2}.}, since some contain degenerate camera motions or large sky regions whose depth ground truth is not reliable.
For the ETH3D dataset, we eliminate those dark sequences (with names containing `\texttt{dark}') to avoid large outliers during comparison, resulting in 50 scenes in total.

\PAR{Baselines and metrics.} We use the same setting as in \cref{subsec:pose} for the VGGT and MegaSAM baselines. Additionally, as a reference, we add monocular metric depth estimation methods including DepthPro~\cite{bochkovskii2024depth} and UniDepth~\cite{piccinelli2025unidepthv2}, where the depth maps are obtained by applying the model on each frame independently.
Following the practice in \cite{wang2025cut3r}, we use the relative absolute depth error (RelAbs), log of depth RMSE error (LogRMSE), and relative depth ratio error ($\delta_{1.25}$).

\begin{table}[!t]
\centering
\footnotesize
\begin{tabular}{l|ccc|ccc}
\toprule
         & \multicolumn{3}{c|}{\textbf{SINTEL}~\cite{butler2012naturalistic}}        & \multicolumn{3}{c}{\textbf{ETH3D}~\cite{schops2019bad}}          \\
         & RelAbs $\downarrow$ & LogRMSE $\downarrow$ & $\delta_{1.25}$ $\uparrow$ & RelAbs $\downarrow$ & LogRMSE $\downarrow$ & $\delta_{1.25}$ $\uparrow$ \\ \midrule
DepthPro~\cite{bochkovskii2024depth} & 0.29   & 0.31    & 62.8            & 0.31   & 0.37    & 64.4            \\
UniDepth~\cite{piccinelli2025unidepthv2} & 0.23   & \cellcolor{second}{0.28}    & \cellcolor{second}{79.8}            & \cellcolor{second}{0.19}   & \cellcolor{second}{0.27}    & \cellcolor{second}{72.1}            \\ \midrule
VGGT~\cite{wang2025vggt}     & \cellcolor{second}{0.22}   & 0.36    & 75.7            & 0.20   & 0.28    & 69.9            \\
MegaSAM~\cite{li2025megasam}  & 0.29   & 0.33    & 67.9            & 0.23   & 0.28    & 64.9            \\ \midrule
\textbf{Ours}     & \cellcolor{first}{0.21}   & \cellcolor{first}{0.27}    & \cellcolor{first}{80.8}            & \cellcolor{first}{0.16}   & \cellcolor{first}{0.22}    & \cellcolor{first}{81.7}            \\ \bottomrule
\end{tabular}
\caption{Depth estimation accuracy measured on \textbf{synthetic and real-world indoor} datasets.}
\label{tab:depth}
\end{table}

\begin{figure}[!t]
\centering
\includegraphics[width=0.9\linewidth]{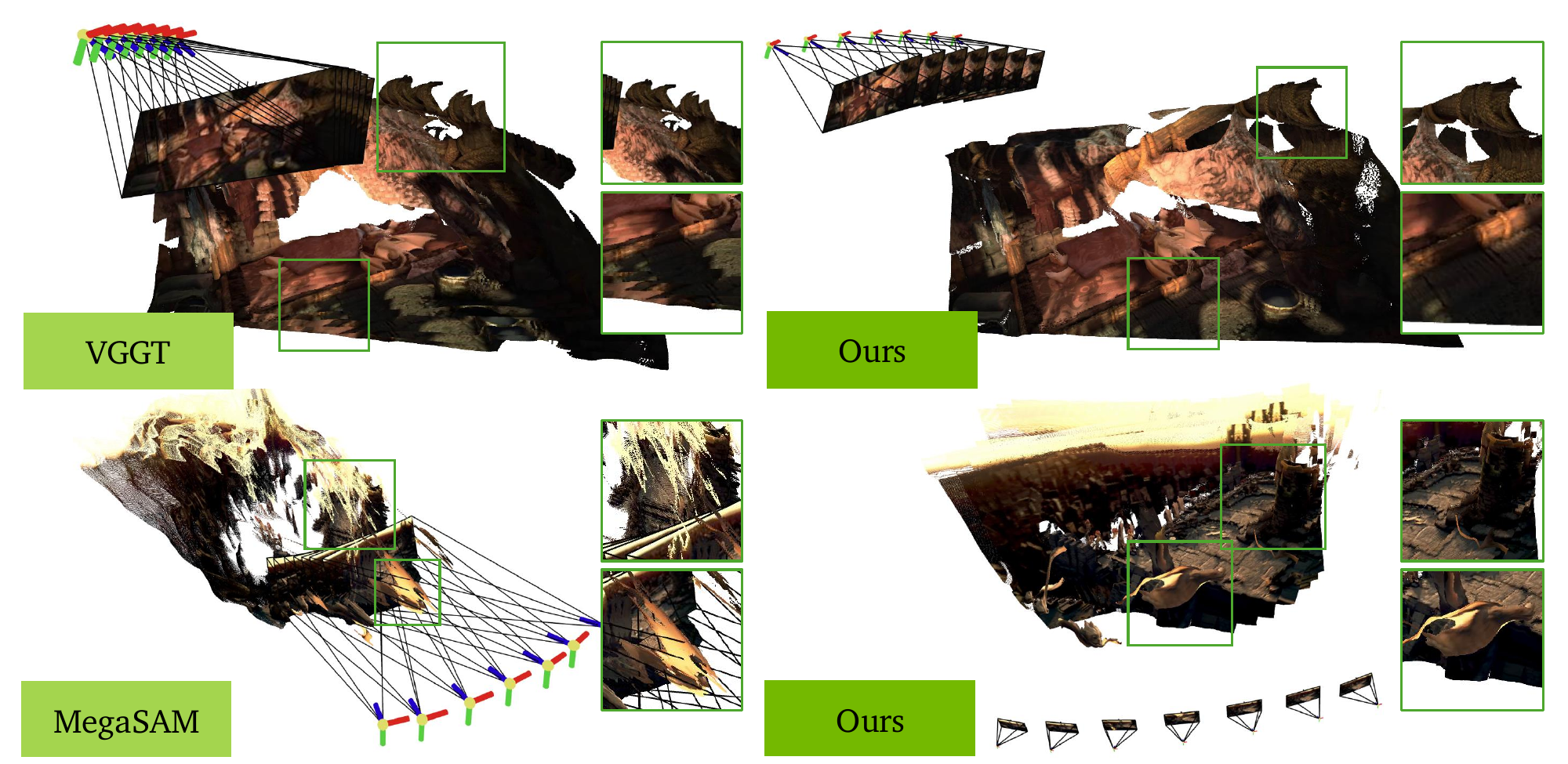}
\caption{\textbf{Qualitative comparisons} of the method output with the baselines on the \textbf{SINTEL} dataset. We subsample the camera frames for clarity only in the visualization.}
\label{fig:sintel_depth}
\end{figure}

\PAR{Results.} Quantitative results are shown in \cref{tab:depth} on the two datasets.
Empirically, we found that although the depth maps coming from the monocular depth estimation methods have better metrics, there is usually non-negligible jittering across frames. \MethodName does not suffer from this issue due to the use of the video depth model.
As shown in \cref{fig:sintel_depth}, after accumulating multiple unprojected point clouds into the 3D world, the baselines have multi-layer artifacts due to the inaccurate depth estimation or camera parameter estimation, and \MethodName is generally robust under these settings.

\subsection{Ablation Study}

In \cref{table:ablation}, we show the effectiveness of different components introduced in \cref{sec:core}.
We use the \textbf{OpenDV} and \textbf{VidBench} datasets as introduced in \cref{subsec:pose_unposed} for evaluation since the videos are more reflective of real-world scenarios.
Experimental results show that adding sparse track terms and dynamic masking improves the robustness of the estimation, while the use of the depth estimation module in the pipeline is able to further improve the accuracy.

\begin{table}[!t]
\centering
\footnotesize
\begin{tabular}{cccc|ccccc}
\toprule
\multirow{2}{*}{$e_\text{dense}$} & \multirow{2}{*}{$e_\text{sparse}$} & \multirow{2}{*}{$e_\text{depth}$} & \multirow{2}{*}{Masking} & \multicolumn{4}{c}{\textbf{OpenDV}~\cite{yang2024generalized}} & \textbf{VidBench~\cite{rockwell2025dynamic}} \\
 &   &  &   & S-ATE ($\times 10^{-2}$) $\downarrow$  & S-RTE ($\times 10^{-4}$) $\downarrow$ & S-RRE ($^\circ$) $\downarrow$ & S-Focal ($^\circ$) $\downarrow$ & Sampson (px) $\downarrow$ \\ \midrule
$\checkmark$ & & & & 1.39  & 4.40 & 0.04  &  5.28  &   1.40    \\
$\checkmark$ &   & $\checkmark$ &   &  1.40    &  4.45 & 0.04  & 5.20  &   1.36      \\
$\checkmark$ & $\checkmark$  & $\checkmark$ &   &  1.35    &  4.21    &  0.04    &  5.00     &    0.84     \\
$\checkmark$ &   & $\checkmark$ &  $\checkmark$ &  1.13    & 4.10     &   0.03   &   4.45    &    0.96     \\ \midrule
$\checkmark$ & $\checkmark$  & $\checkmark$ & $\checkmark$  &  \textbf{1.05}    &   \textbf{3.80}   &  \textbf{0.03}    &   \textbf{4.26}    &    \textbf{0.83}     \\ \bottomrule
\end{tabular}
\caption{\textbf{Ablation study} on the effectiveness of different components in \MethodName.}
\label{table:ablation}
\end{table}

%% file: sections/05-dataset.tex
\section{Dataset Release}

\PAR{Overview.}
To address the scarcity of high-quality, diverse, and large-scale datasets for 3D geometric perception in unconstrained environments, and to facilitate future research in this field and its downstream applications, we introduce and release three new datasets annotated with \MethodName's camera poses and geometric information. These datasets span a wide range of video sources and content, providing high diversity for robust visual learning.
These include:
\begin{itemize}[leftmargin=*, itemsep=0pt, topsep=0pt]
    \item \textbf{Dynpose-100K++}: Dynpose-100K~\cite{rockwell2025dynamic} is a dataset containing $\sim$100K real-world videos gathered and filtered from the Internet. The videos are originally taken from the PANDA-70M~\cite{chen2024panda} dataset and several filters have been used on top to filter out sequences that are not suitable for camera pose estimation. However, the dataset has its pose annotated with a Structure-from-Motion pipeline whose camera poses are provided at a lower framerate (12FPS) than the actual video. Furthermore, no per-frame geometry is provided, making verification and evaluation of the quality challenging. We hence re-annotate the dataset (hence the name `++') using our approach, resulting in 99,501 videos with 15.7M frames spanning $\sim$150 hours in total.

    \item \textbf{Wild-SDG-1M}: Recently video diffusion models~\cite{agarwal2025cosmos,wan2025wan} have demonstrated impressive quality given text prompts. Compared to real-world videos, the videos generated with state-of-the-art diffusion models, with well chosen prompts, are often  clear and of high quality, reducing the need for further filtering. We sampled $\sim$1 million videos from the video diffusion models using our in-house curated and balanced text prompts, and annotated all the sampled frames using \MethodName, resulting in $\sim$78 million frames in total.
    \item \textbf{Web360}: Web360 is a relatively small-scale dataset containing 360-degree panorama videos curated by the authors of \cite{wang2024360dvd}. The dataset contains approximately 2,000 videos in ERP format from the Internet and games. We release per-frame camera poses and distance maps for this dataset.
\end{itemize}

\PAR{Impact.}
These newly released datasets offer a valuable resource for downstream developers --  dataset scale and diversity—spanning real-world dynamic internet videos, synthetic environments, and specialized panoramic content—make them well-suited for training and evaluating 3D geometric perception models under a variety of challenging conditions.
We hope that this release contributes to advancing downstream real-world applications.

\begin{figure}
\centering
\includegraphics[width=0.9\linewidth]{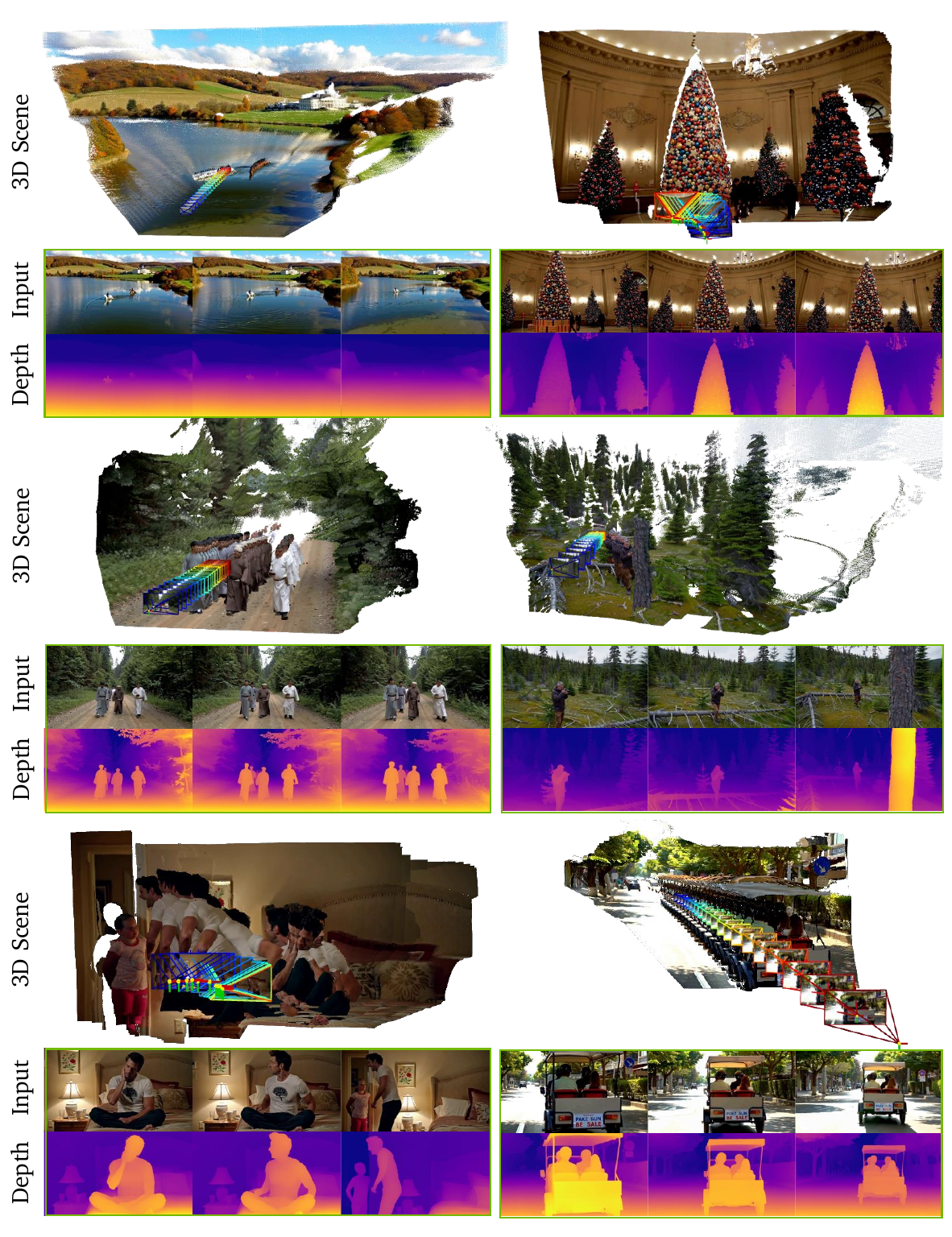}
\caption{Qualitative demonstration of \textbf{Wild-SDG-1M} dataset annotated using \MethodName. For the upper 3D scene, we accumulate the point clouds for both the static and dynamic parts of the scene, while the lower rows show the input samples from the videos and the corresponding estimated depth maps.}
\label{fig:qual_sdg}
\end{figure}

\begin{figure}
\centering
\includegraphics[width=0.9\linewidth]{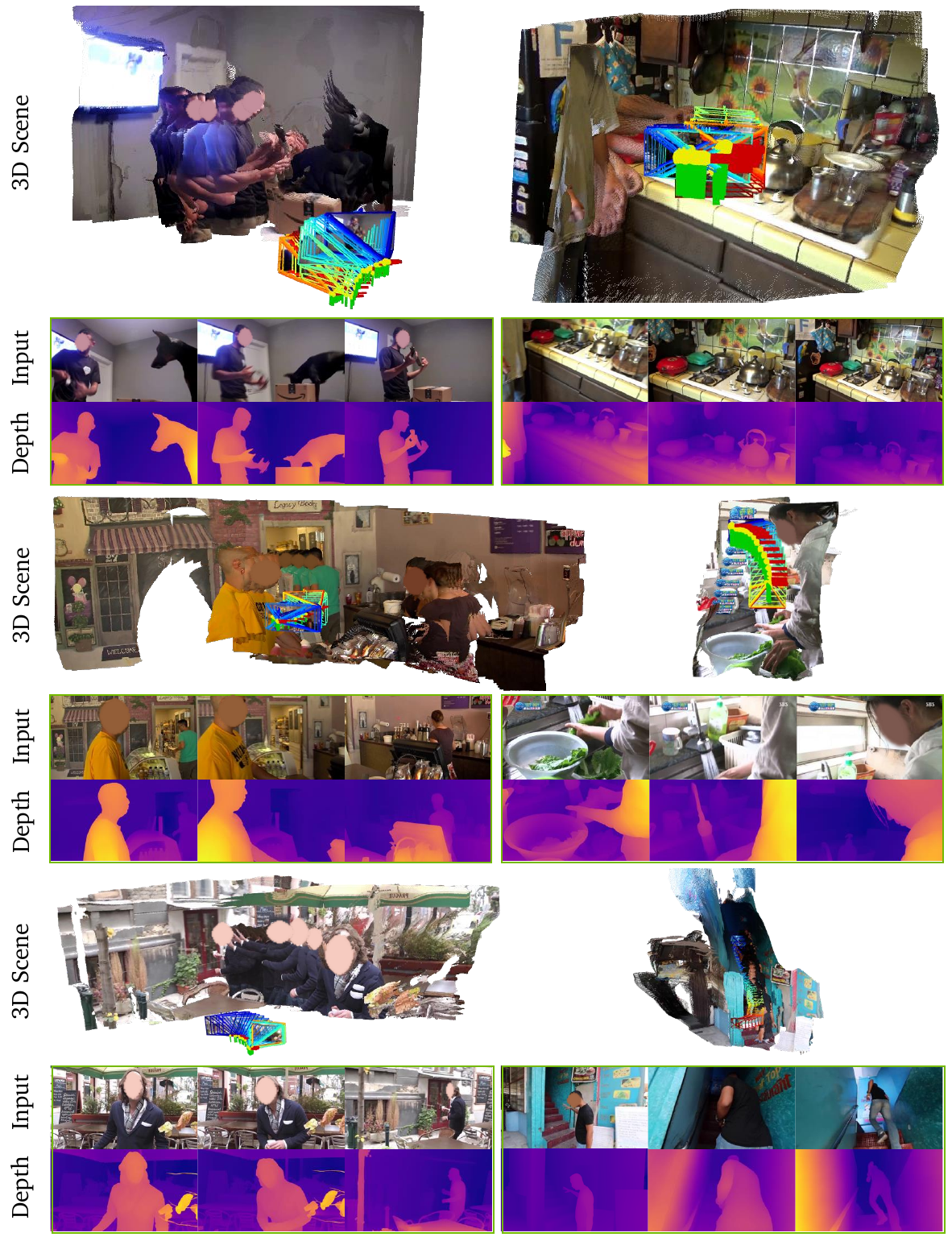}
\caption{Qualitative demonstration of \textbf{DynPose-100K++} dataset annotated using \MethodName.}
\label{fig:qual_dynpose}
\end{figure}

\begin{figure}
\centering
\includegraphics[width=0.9\linewidth]{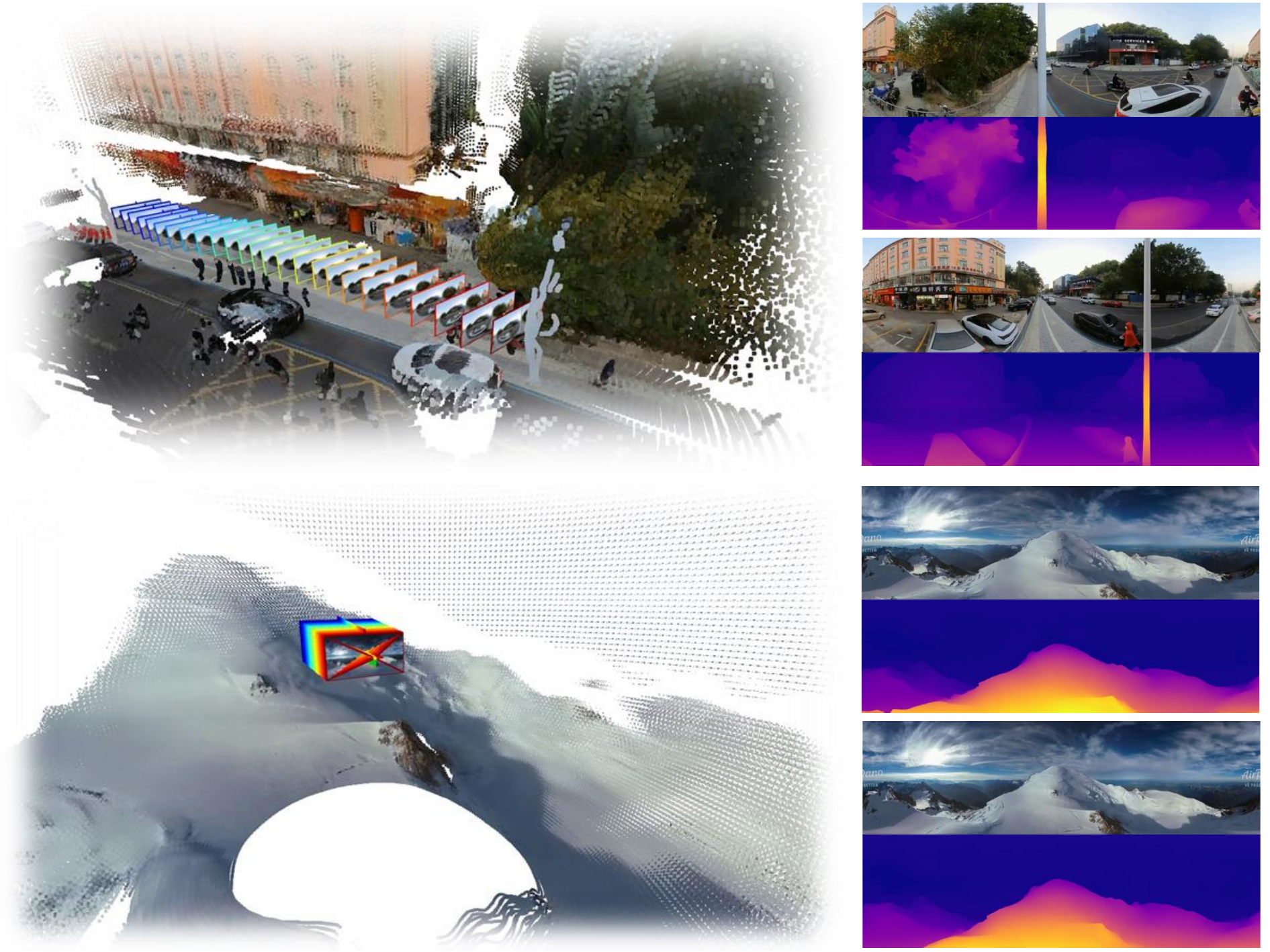}
\caption{Qualitative demonstration of \textbf{Web360} dataset annotated using \MethodName.}
\label{fig:qual_360}
\end{figure}

\PAR{Qualitative Results.}
To visually demonstrate the consistency and robustness of \MethodName's annotations across highly diverse video types, we present a selection of qualitative results from our newly released datasets in \cref{fig:qual_sdg,fig:qual_dynpose,fig:qual_360}. Note that in the examples we showcase the quality that \MethodName consistently achieves, particularly in conditions where other methods may struggle or yield incomplete estimations.

%% file: sections/06-conclusion.tex
\section{Conclusion}

In this work, we present \MethodName, a video pose estimation engine that estimates camera poses, intrinsics, and depth maps from videos. The system is built upon a bundle adjustment framework that integrates both dense and sparse constraints, leveraging the strengths of both optical flow and keypoint tracking. We also introduce a depth alignment strategy to ensure consistent depth maps across frames.
Our method is benchmarked on a variety of datasets, including both static/dynamic and indoor/outdoor scenes, and shows superior performance compared to existing methods.

In practice, \MethodName has seen wide adoption across downstream applications with notable impact. It has been used to annotate training data and produce conditional buffers for world generation in Gen3C~\cite{ren2025gen3c} and Cosmos~\cite{agarwal2025cosmos}. The annotated datasets have also supported BTimer~\cite{liang2024feed} in reconstructing 3DGS in a feed-forward manner, improving its robustness to diverse inputs. We hope our released dataset will continue to drive advances in 3D geometric perception and related fields.